\documentclass[journal, 9pt, a4paper]{IEEEtran}
\usepackage[utf8]{inputenc}
\usepackage{
    float,
    marvosym,
    graphicx,
    siunitx,
    amsmath,
    amsfonts,
    mhsetup,
    mathtools,
    csquotes
}
\usepackage{todonotes} 
\usepackage{hyperref}
\usepackage[english]{babel}
\usepackage[backend=biber, sortcites=true, sorting=none]{biblatex}
\usepackage{multicol}
\usepackage{orcidlink}
\usepackage{float}
\usepackage{dblfloatfix}
\usepackage{placeins}
\hypersetup{hidelinks, colorlinks=false,}

\DeclareMathOperator{\atantwo}{atan2}

\setuptodonotes{inline,color=blue!25,size=\small}

\begin{document}
    \title{A Modular Architecture Design for Autonomous Driving Racing in
    Controlled Environments}
    \author{Brais Fontan-Costas\orcidlink{0000-0001-7433-7608}, M. Diaz-Cacho\orcidlink{0000-0002-2784-1374},
    Ruben Fernandez-Boullon\orcidlink{0009-0006-9695-4686}, Manuel Alonso-Carracedo\orcidlink{0009-0001-5037-5826},
    Javier Perez-Robles\orcidlink{0009-0006-4032-8301}}

    \maketitle


    %
    %
    %
    %
    \begin{abstract}
        This paper presents a modular autonomous driving architecture for Formula Student Driverless competition vehicles operating in closed-circuit environments. The perception module employs YOLOv11 for real-time traffic cone detection, achieving 0.93 mAP@0.5 on the FSOCO dataset, combined with neural stereo depth estimation from a ZED 2i camera for 3D cone localization with sub-0.5 m median error at distances up to 7 m. State estimation fuses RTK-GNSS positioning and IMU measurements through an Extended Kalman Filter (EKF) based on a kinematic bicycle model, achieving centimeter-level localization accuracy with a 12 cm improvement over raw GNSS. Path planning computes the racing line via cubic spline interpolation on ordered track boundaries and assigns speed profiles constrained by curvature and vehicle dynamics. A regulated pure pursuit controller tracks the planned trajectory with a dynamic lookahead parameterized by speed error. The complete pipeline is implemented as a modular ROS 2 architecture on an NVIDIA Jetson Orin NX platform, with each subsystem deployed as independent nodes communicating through a dual-computer configuration. Experimental validation combines real-world sensor evaluation with simulation-based end-to-end testing, where realistic sensor error distributions are injected to assess system-level performance under representative conditions.
    \end{abstract}
    %
    %
    %
    %
    \section{Introduction}
    Autonomous vehicle systems in controlled environments present significant
    challenges in integrating multiple subsystems for real-time navigation and
    decision-making. The development of modular architectures that effectively combine
    perception, localization, path planning, and control systems represents a
    critical area of research in autonomous driving technology. This work presents
    a comprehensive framework for the connectivity and allocation of responsibilities
    within an autonomous driving architecture, focusing on precise operation in
    closed-circuit scenarios. The approach defines four primary modules:
    perception, localization and mapping, trajectory planning, and control. It
    also describes their interconnection through a communication pipeline. The paper
    also reviews the current state of the art, analyzes the main technological
    advances, and justifies the design choices made to address the scientific and
    engineering challenges faced by autonomous vehicles in constrained,
    competitive or experimental environments.


    %
    %
    \section{Autonomous Vehicle System Overview}
\begin{figure}
    \centering
    \includegraphics[width=1\linewidth]{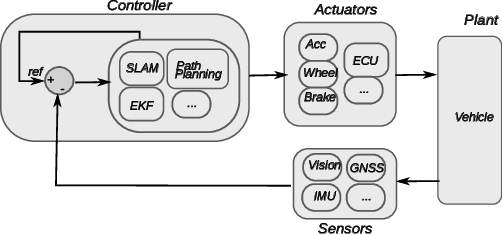}
    \caption{Autonomous Vehicle Control System}
    \label{ControlSystem}
\end{figure}

The architecture of an autonomous vehicle is a particular implementation of a more generic control system. Therefore, the classic components of sensors, actuators, controller, and plant must be reinterpreted and mapped to the autonomous driving ecosystem. Figure \ref{ControlSystem} shows a simplified version of this projection. The sensors are mainly a vision system, an IMU (Inertial Measurement Unit), and a positioning system. The actuators are concentrated into an ECU (Electronic Control Unit) system, which translates the commands received by the controller into direct actions on the steering wheel, accelerator, and brake systems. The controller is the autonomous system itself, which computes at each iteration a new reference (velocity) based on a path planning algorithm and compares it with the received position, video, and IMU data, constituting the SLAM (Simultaneous Localization and Mapping) system.

    \subsection{Hardware Components}
    \subsubsection{Control System}
    The Control System of the Autonomous System (AS) is suitable to be implemented entirely on an On-Board-Computer. The On-Board-Computer has to have quite high computing capabilities in order to solve in real time several complex calculations like integration of positioning, IMU and Vision systems into an Extended Kalman Filter (EKF), and the use of several AI (Artificial Intelligence) algorithms. 

    \subsubsection{Sensors}
    The data generated by the sensors serves as input to the Control System in order to be compared with the desired reference. The sensors are installed to determine the behavior of the vehicle, and particularly its absolute or relative position within a defined area. 
    \begin{itemize}
        \item Position sensors. These sensors help the vehicle understand its coordinates and orientation relative to the environment. The most used ones are GNSS (Global Navigation Satelite Systems), LiDAR (Light Detection and Ranging), GSS (Ground Speed Sensors), odometry sensors (encoders), IMU among others.
        \item Vision sensors. Image processing has become an important source of data for obstacle detection and environmental interpretation. Mono and stereo cameras complemented with image processing software generates useful control input data.
        \item Odometry. Motor position encoders are suitable for accurate short-distance travel. Steering angle and wheel encoders are examples for that.
        \item Cinematic. Torque and acceleration data are provided by IMU in order to detect rapid behavior changes.
    \end{itemize}

    \subsubsection{Actuators}

    Actuators are mainly concentrated into three components: the throttle actuator, the steering system and the brake system. Usually, the output of the Control System is an acceleration vector. Nevertheless, this acceleration is translated into the steering system, the linear acceleration and the brake system. Usually, in a vehicle, these actuators are connected using field bus technologies such as CAN bus, where the communication is controlled by an Electronic Control Unit (ECU).
    
    \subsubsection{In Vehicle Communications}
    Data transmission between the components of the vehicle is based on the Controller Area
    Network (CAN) protocol \cite{iso11898}. Usually, the interface adheres to 
    a specific software specification depending on the vehicle.

    %
    %
    \subsection{Software}
    The software components bind all the data generated and needed by the hardware components together. The On-Board computer that hosts the Control System has to process the inputs that come from the sensors, generate the outputs, and send them to the actuators. Therefore, the On-Board computer hosts the most important part of the software components.
    
    \section{Implementation}

    The presented AS is developed to compete in the Formula Student Driverless competition \cite{FSUK}, and was deployed at Formula Student UK (FSUK) at Silverstone Circuit. The competition organizers provide an electric car with all actuators integrated in a Vehicle Control Unit (VCU) connected to an In-Car-PC equipped with Ethernet and CANBus interfaces. The interface adheres to the AS-DV software specification defined for the competition \cite{ads_dv_spec}, categorizing messages into control commands, sensor feedback, and mission status flags. The Control System is hosted by an On-Board computer; therefore, this architecture is called a dual-computer architecture (On-Board-Computer and In-Car-PC).
    The circuit is delimited by two sets of cones: blue cones for left track limits and yellow cones for right track limits, with large orange cones marking the start and finish lines. This requires an object detection system to be integrated into the AS.

    \subsection{Selected Hardware Components}

The main hardware components are the On-board computer that hosts the Control System and the position and vision sensors. As position sensor, the authors propose a Real-Time-Kinematic (RTK) system in order to improve classical GNSS data and as vision sensor a stereo camera in order to facilitate the calculation of depth and positioning relative to the location of the cones.

\begin{figure}[!h]
        \centering
        \includegraphics[width=\linewidth]{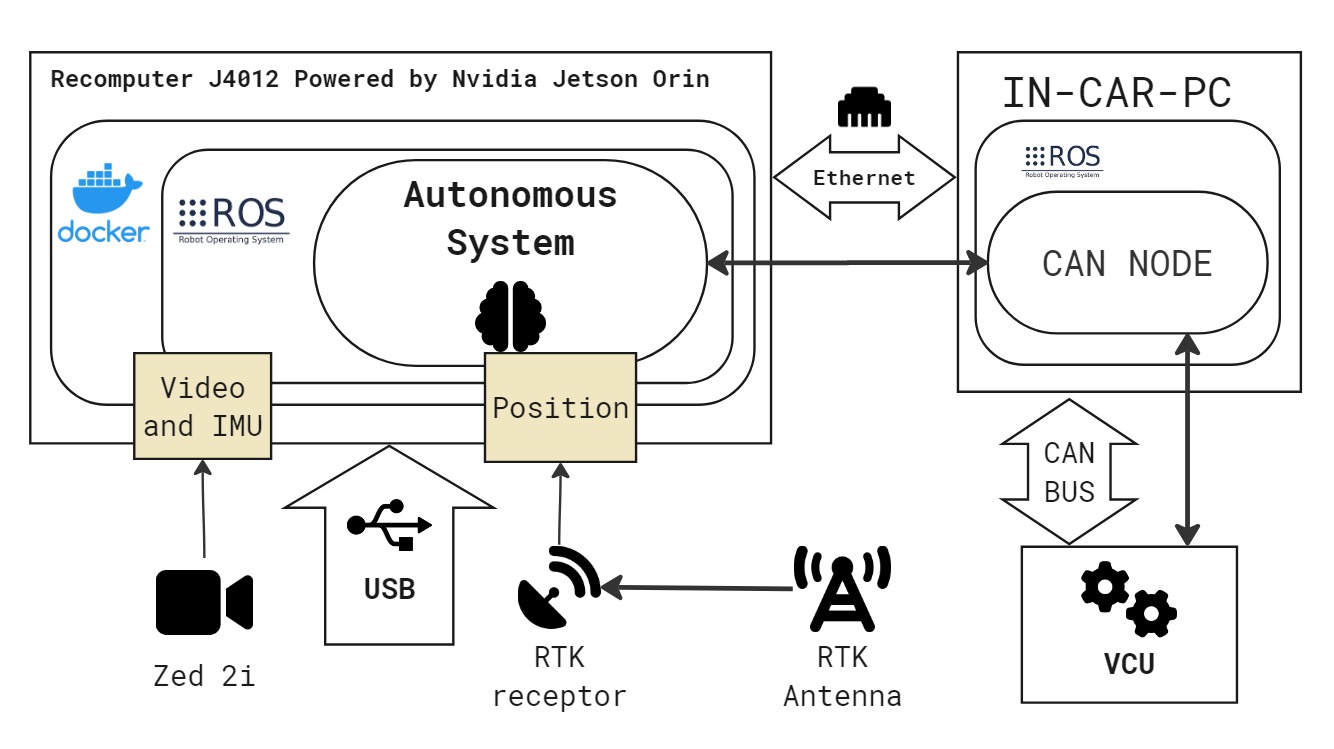}
        \caption{Hardware Architecture}
        \label{hardware}
    \end{figure}

The whole hardware architecture (dual-computer architecture) consists of five primary components: a ZED 2i stereo camera, a simpleRTK2B receiver, a reComputer-based On-Board-Computer (primary AS host), a dedicated In-Car-PC (secondary computer for CAN communication), and the VCU. The On-Board-Computer executes all perception, planning, and control modules, while the in-car PC handles CAN Bus communication and the interface with the Vehicle Control Unit. This separation ensures that computationally intensive algorithms run independently from real-time communication interfaces. The hardware components and the connectivity are shown in figure \ref{hardware}.
 
    \subsubsection{On-board computer} The Control System of the AS runs on a reComputer J4012 powered by the NVIDIA\textregistered
    Jetson Orin NX module \cite{nvidia_orin}. This System-on-Module (SoM) features
    an integrated Ampere GPU with 1024 CUDA cores and dual deep learning accelerators,
    delivering up to 100 TOPS of AI performance for the vision pipeline. The system
    operates on Ubuntu 20.04 with ROS 2 Galactic, utilizing the architecture
    described by Macenski et al. \cite{macenski2022ros2}. The middleware is
    virtualized via Docker to ensure compatibility, while the in-car PC runs Ubuntu
    22.04 with ROS 2 Humble.

    \subsubsection{Real-time kinematic positioning} The simpleRTK2B module from ArduSimple \cite{ardusimple_rtk} provides centimeter-level
    GNSS accuracy. It utilizes the u-blox ZED-F9P multiband GNSS engine to
    receive corrections from a fixed base station up to 11 km away. The rover board
    computes Real-Time Kinematic (RTK) solutions at a rate of 10 Hz, providing
    the high-precision absolute positioning required for the Extended Kalman
    filter updates.

    \subsubsection{Stereo camera} The ZED 2i \cite{zed2i_datasheet} captures 1920x1080 stereo images at 30fps
    with a 120º field of view. It integrates a 9-DOF IMU (accelerometer, gyroscope,
    and magnetometer) running at 400 Hz, which is factory-calibrated and
    synchronized with the video stream. Its IP66-rated enclosure ensures reliability
    during adverse weather conditions typically encountered in competition.

    \subsection{Software components}

    The On-board computer hosts the whole Control System of the AS. Communications with the other components are done by the parametrization and correct use of the operational drivers of the sensors and actuators. 

\begin{figure}[!h]
        \centering
        \includegraphics[width=\linewidth]{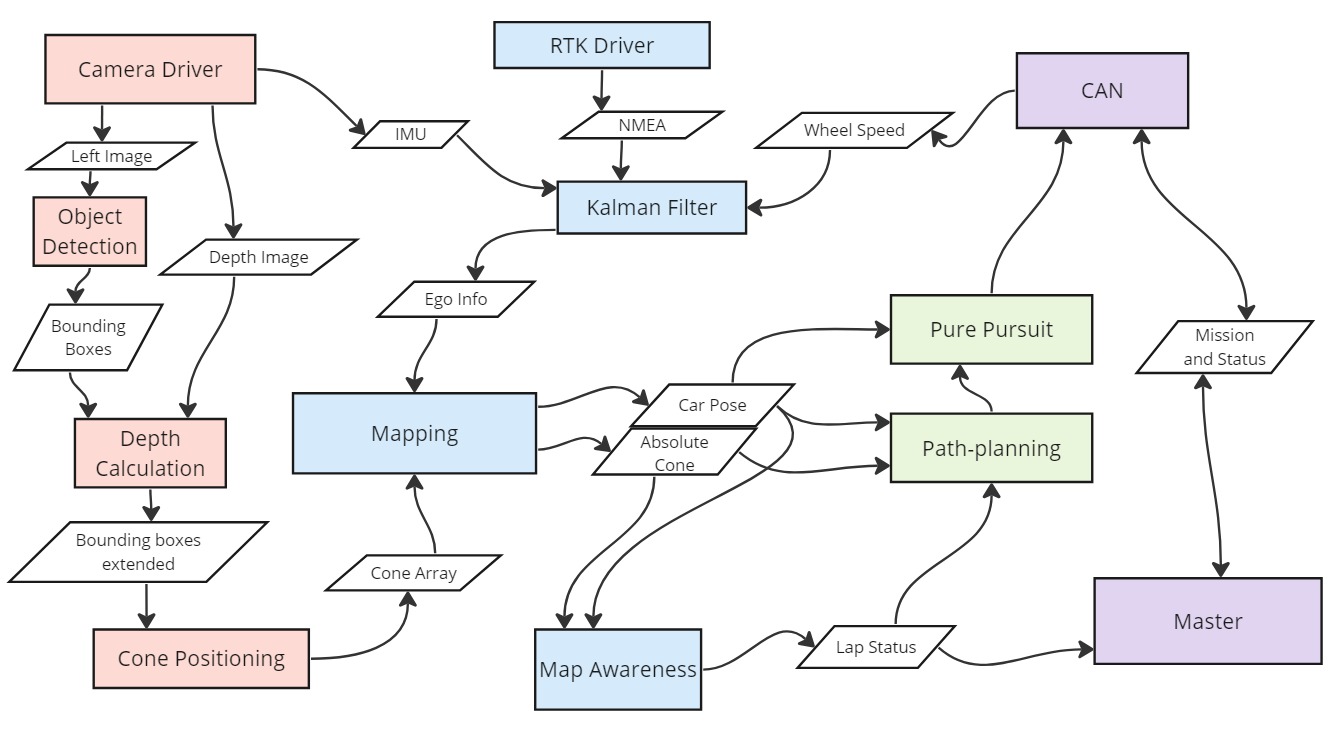}
        \caption{Software Architecture}
        \label{Software_diagram}
    \end{figure}

    The AS is developed on ROS2 \cite{ros}, a software framework that provides utilities
    for complex robotic machines. Each module is implemented via a ROS2 package
    and each submodule is implemented via ROS2 nodes as illustrated in Figure
    \ref{Software_diagram}.

    \subsection{Computer Vision}

    The main objective of the computer vision module is to take camera images
    and process them in order to detect the cones that delimit the track. The object
    detection and depth calculation nodes take left and right images from the camera
    driver. A computer vision model is used to detect the cones delimiting the
    track, while a depth calculating algorithm calculates a depth image. Using
    both the detections and the depth information, the positioning node calculates
    a bird's eye view positioning of the cones relative to the car. The nodes
    that are part of the computer vision module are marked in red in the figure
    \ref{Software_diagram}.

    \subsubsection{Depth calculation}
    Stereolabs offers as part of their SDK for the ZED a depth calculation
    functionality \cite{zed2}. It allows a variety of depth calculation modes,
    that change the performance and speed of the calculation. During our experiments,
    the best results were obtained by using the Neural mode. This mode uses an end
    to end AI powered calculation. This calculation is very computer intensive
    and uses the API for parallel computing CUDA to compute the depth in the reComputer. The use of a visual tracker,
    combined with the mapping algorithm mitigates the depth error.

    \subsubsection{Object Detector}
 For object detection (cones)  the version 11 of the open source YOLO product from Ultralytics is selected \cite{terven2023comprehensive}.

    YOLOv11 \cite{yolov8} was released in September 2024 by Ultralytics
    \cite{ultralytics}. There are some pre-prints that analyse the performance of
    the YOLOv11 model \cite{terven2023comprehensive}. In these pre-prints the
    YOLOv11 model performed better than its predecessors in speed and accuracy. For
    these reasons, YOLOv11 was chosen as the detection model.

    Ultralytics offers different pretrained models to use as training baseline. The
    YOLOv11 model chosen was the YOLOv11s object detection model, since in our
    experiments the accuracy difference between the 's' and 'm' models was not
    significant, while the speed difference was noticeable.

    THe YOLO solution has to be trained using a dataset. The FSOCO dataset was used to train the model \cite{fsoco_2022}. 
    The best results were obtained at around 40 iterations and further training seemed
    to cause overfitting. To search the best combination for the 30
    hyperparameters a genetic algorithm was used to optimize a fitness function
    defined as follows:

    \begin{equation}
    \mathcal{F} = 0.1 \cdot \mathrm{Recall} + 0.1 \cdot \mathrm{mAP}_{50} + 0.8 \cdot \mathrm{mAP}_{50:95}
    \label{eq:fitness}
    \end{equation}

    The fitness function defined in Equation \ref{eq:fitness} assigns the dominant weight (0.8) to mAP\textsubscript{50:95}, which evaluates detection accuracy across IoU thresholds from 0.50 to 0.95 in steps of 0.05, penalizing detections with imprecise bounding boxes. Recall and mAP\textsubscript{50} each contribute 0.1, ensuring that the optimization does not sacrifice detection completeness or baseline localization quality. This weighting reflects the priority of tight spatial accuracy over mere detection presence, which is critical for downstream cone positioning: a detection that loosely overlaps a cone may count as correct at IoU 0.50 but fail at stricter thresholds, propagating localization error into the mapping and path planning modules. The genetic algorithm optimizes this function using a 90 percent mutation probability with 0.04 variation to generate new candidates from the best-performing parents of previous generations.

    \subsubsection{Object tracker}
    A multi object tracker (MOT) is an algorithm that allows recovering the identity
    information of several detected object across frames \cite{luo2021multiple}.
    In the presented design an object tracker is used to maintain the cone
    identities across time. Since the mapping algorithm will also provide a way to
    identify the cones, there is no need to have a highly precise tracker, therefore
    it was chosen the Intersection-Over-Union (IOU).

    The method assumes that an object detector produces a detection per frame for
    every object to be tracked. It also assumes that consecutive detections of
    an object have a high overlap IOU, which is true in sufficiently high frame rates.
    The IOU measure can be defined using the equation \ref{IOU_equation}.

    \begin{equation}
        IOU(a,b) = \frac{ Area(a)\bigcap Area(b)}{Area(a) \bigcup Area(b)}\label{IOU_equation}
    \end{equation}

    When both requirements are met, tracking with this method can be done
    without using any image information. The tracker continues the identity by associating
    the detection with the highest IOU to the last detection in the previous frame
    if a certain threshold $\sigma_{IOU}$ is met \cite{1517Bochinski2017}. All
    detections not assigned to an existing track will start a new one. All tracks
    without an assigned detection will end.

    \subsubsection{Cone positioning}

    The 2D positioning is calculated using both the detected images and the
    depth information. The position was modeled using a distortion-free projective
    transformation given a pinhole camera model following Equation \ref{positioning_equation}.
    \begin{equation}
        s\cdot p = A[R|t]P_{w}\label{positioning_equation}
    \end{equation}
    Where $P_{w}$ is a 3D point expressed with respect to world coordinates, $p$
    is a 2D pixel in the image plane, $A$ is the camera intrinsic matrix, $R$ and
    $t$ are the rotation and translation that describe the change of world coordinates
    to camera coordinate systems, and $s$ is the projective transformation's
    arbitrary scaling and not part of the camera model \cite{opencv_recons}. The
    camera intrinsic matrix $A$ is composed of the lengths expressed in pixel units ($f_x, f_y$)
    and the principal point ($c_x,c_y$) following the equation \ref{camera_matrix}.
    \begin{equation}
        A =
        \begin{bmatrix}
            f_{x} & 0     & c_{x} \\
            0     & f_{y} & c_{y} \\
            0     & 0     & 1
        \end{bmatrix}
        \label{camera_matrix}
    \end{equation}
    The world coordinates are equal to the camera coordinates since the
    translation and rotation will be calculated in the mapping node. Therefore,
    the rotation and translation matrix is the identity $[R|t]=[I|0]$. This model
    allows to calculate the image projected given the real coordinates. The
    depth information can be used to calculate the z given the camera intrinsic
    parameters and the pixel position using similar triangles.

    \subsection{Positioning and mapping}

    The Positioning and Mapping module captures sensor data to determine accurate
    location and orientation. It consists of the submodules Camera Driver, RTK Driver,
    Extended Kalman Filter, Mapping, and Map Awareness. RTK and IMU data are the
    main inputs, which are combined using a Extended Kalman Filter to achieve
    high accuracy. The Mapping submodule creates a global map of the circuit by combining
    location and computer vision data, assigning a linear order to cones and
    publishing the car's position and map information. Additionally, the map
    awareness node provides information about the car's position on the track.

    \subsubsection{Extended Kalman Filter}

    A Kalman filter \cite{kalman_filter} is an algorithm that uses a series of measurements
    observed over time to produce estimates of unknown variables. The algorithm
    works in a two-phase process: predict and update. In the prediction phase, the
    filter estimates the current state variables and uncertainties. In the
    update phase, these estimates are corrected using a weighted average of the new
    measurements, prioritizing those with greater certainty.

    In the context of autonomous racing, state estimation typically relies on fusing
    high-frequency ego sensors (IMU, odometry) with positioning systems (GNSS-RTK).
    For this architecture, we employ the \textit{Kinematic Bicycle/Tricycle Models}  as the
    dynamical system foundation. These models are widely validated in closed
    circuit racing operations \cite{Kabzan2019AMZ}, \cite{Raji_2023} for their optimal balance
    between computational load and prediction accuracy.

    While dynamic models account for complex tire-slip mechanics, the kinematic approach
    fused with GNSS updates has been proven to maintain decimeter-level accuracy
    ($<15$ cm) in dynamic environments \cite{Albarakati2022Experimental}.
    Furthermore, recent experimental comparisons demonstrate that EKF fusion using
    this modeling approach can improve localization accuracy by approximately 32\%
    compared to raw sensor data \cite{Li2025Adaptive}, therefore it's selected for
    achieving the 1-2 cm target accuracy when combined with RTK corrections.

\begin{figure}[H]
        \centering
        \includegraphics[width=0.5\linewidth]{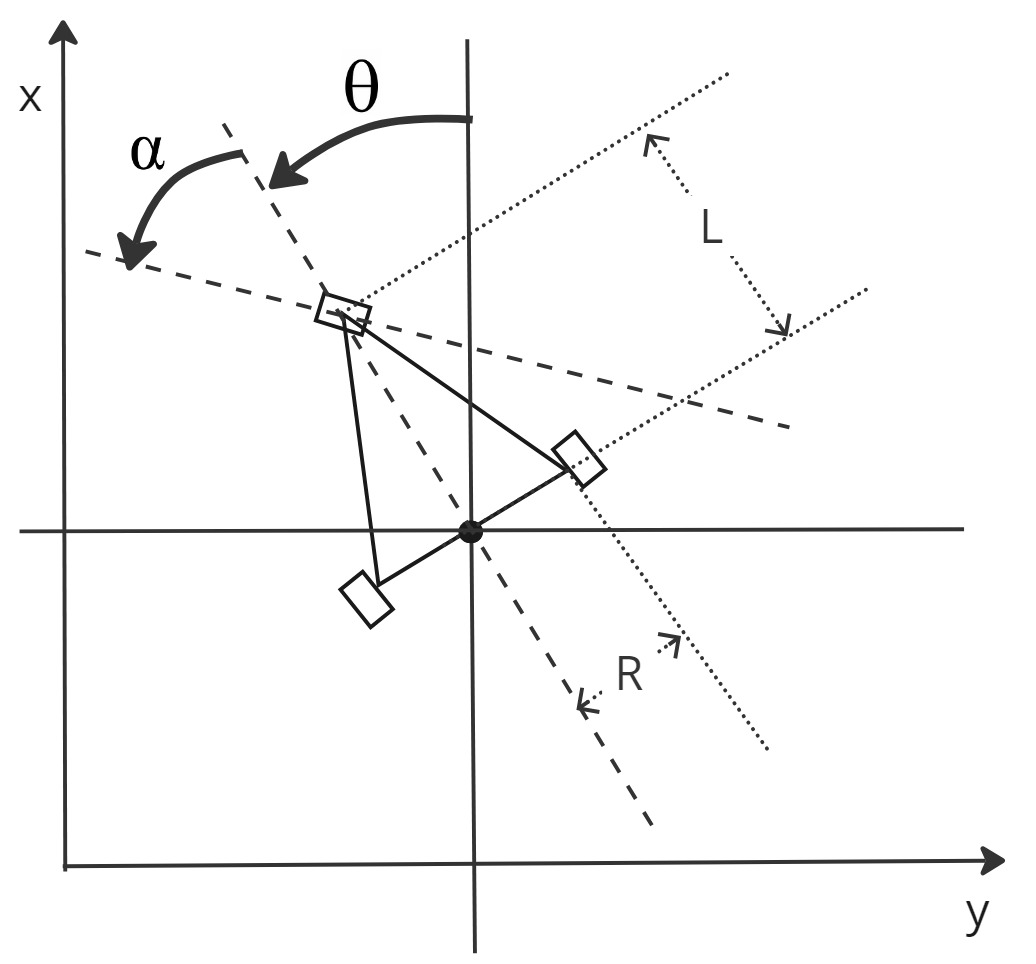}
        \caption{Kinematic Bicycle Model (Tricycle representation)}
        \label{tricycle}
    \end{figure}

    The control system outputs are the longitudinal acceleration $a$ and the steering
    angle $\alpha$, and the inputs are the GNSS coordinates
    $(x, y)$, the magnetometer heading angle $\theta$, the vehicle speed $v$ (median
    of rear wheel speeds), and the IMU angular velocity $\omega$, as shown in
    Figure \ref{tricycle}.

    To model the system, it is required to represent a linear system using a
    propagation model (equation \ref{propagation model}) and a measuring model (equation
    \ref{Measuring model}).


    \begin{align}
        \vec{x}(k+1) & = A\vec{x}(k) + B\vec{u}(k) + \vec{\varepsilon}_1(k)\label{propagation model} \\
        \vec{y}(k)   & = C\vec{x}(k) + \vec{\varepsilon}_2(k) \label{Measuring model}
    \end{align}

    Where $k$ is the current instant, $\vec{x}$ is the internal state vector,
    $\vec{y}$ is the input vector, $\vec{u}$ is the output vector, $A$, $B$ and $C$
    are the model matrices, and $\vec{\varepsilon_1}$ and $\vec{\varepsilon_2}$ are the considered
    errors. The internal state vector is composed of the $x$ and $y$ coordinates,
    the turning angle $\theta$ and the vehicle speed $\nu$.

    Our approach used a uniformly accelerated rectilinear motion (UARM) to
    calculate $x(k+1)$, $y(k+1)$ and $\theta(k+1)$ (equations
    \ref{linear x acceleration}, \ref{linear y acceleration} and \ref{angular
    acceleration}). And lastly for the speed $\nu(k+1)$ we simply develop the acceleration
    formula (equation \ref{speed acceleration}).
    \begin{eqnarray}
        x(k+1) &\simeq& x(k) + \nu_x(k)t_s + \frac{1}{2}a_x(k)t_s^2 \label{linear
        x acceleration}\\
        y(k+1) &\simeq& y(k) + \nu_y(k)t_s + \frac{1}{2}a_y(k)t_s^2
        \label{linear y acceleration}\\
        \theta(k+1) &\simeq& \theta(k) +\dot{\theta}(k)t_s+\frac{1}{2}\ddot{\theta}(k)t_s^2
        \label{angular acceleration}\\
        \nu(k+1) &\simeq& \nu(k) + a(k)t_s \label{speed acceleration}
    \end{eqnarray}
    Where $t_{s}$ is the sampling period. With a sampling period small enough, the
    last component of equations \ref{linear x acceleration}, \ref{linear y
    acceleration} and \ref{angular acceleration} will be much smaller than the other
    components, so it can be ignored to simplify the model. In the design,
    $t_{s}$ will be 0.1 seconds, because the RTK has a maximum refresh rate of 10Hz.
    Developing the above equations, the propagation model will follow the equation
    \ref{propagation matrix}.

    \begin{equation}
        \begin{bmatrix}
            x      \\
            y      \\
            \theta \\
            \nu    \\
        \end{bmatrix}
        (k+1) =
        \begin{bmatrix}
            x(k) - t_{s}\nu(k)\sin{\theta}(k)               \\
            y(k) + t_{s}\nu(k)\cos{\theta}(k)               \\
            \theta (k) + \frac{t_s}{l}\tan(\alpha(k))\nu(k) \\
            \nu(k) + a(k)t_{s}
        \end{bmatrix}
        \label{propagation matrix}
    \end{equation}

    The terms of equation \ref{propagation matrix} are:
    \begin{itemize}
        \item \textbf{$x(k)$, $y(k)$}: Vehicle position coordinates at time step
            $k$. These denote the current location of the vehicle in the global
            (or chosen local) reference frame.

        \item \textbf{$\theta(k)$}: Heading angle (orientation) of the vehicle at
            step $k$, typically in radians.

        \item \textbf{$\nu(k)$}: Longitudinal velocity (speed) of the vehicle at time
            $k$.

        \item \textbf{$t_{s}$}: Sampling interval, i.e., the time difference
            between steps $k$ and $k+1$ (in seconds).

        \item \textbf{$\alpha(k)$}: Steering angle at time step $k$,
            representing the front wheel's angular deviation from the forward
            direction.

        \item \textbf{$l$}: Wheelbase, that is, the distance between the front and
            rear axles of the vehicle.

        \item \textbf{$a(k)$}: Longitudinal acceleration applied at time $k$.
    \end{itemize}

    Each row of the propagation equation gives the predicted value at the next step
    ($k+1$), given the current state and applied control input:
    \begin{itemize}
        \item $x(k+1)$: Next x-position, determined by forward motion and heading.

        \item $y(k+1)$: Next y-position, following a similar principle.

        \item $\theta(k+1)$: Updated heading angle, proportional to current turning
            rate.

        \item $\nu(k+1)$: Updated speed, incremented by the product of acceleration
            and sampling time.
    \end{itemize}

Equation \ref{propagation matrix} can be reformulated in equation \ref{propagation formula} in order to match the state equations of the model (equations \ref{propagation model} and \ref{Measuring model}) and identify the model matrices.
    \begin{align}
        \begin{bmatrix}x\\ y\\ \theta\\ \nu\\\end{bmatrix} (k+1) = \nonumber                                                                                                                                                                                                                                                                                                           \\
        \begin{bmatrix}1&0&0&-t_{s}\sin{\theta}(k) \\ 0&1&0&t_{s}\cos{\theta}(k) \\ 0&0&1&\frac{t_s}{l}\tan{\alpha}(k) \\ 0&0&0&1\end{bmatrix} \begin{bmatrix}x\\ y\\ \theta\\ \nu\end{bmatrix} (k) + \begin{bmatrix}0\\ 0\\ 0\\ t_{s}\end{bmatrix} \begin{bmatrix}a\end{bmatrix} (k) + \begin{bmatrix}\varepsilon_{1_{x}}\\ \varepsilon_{1_{y}}\\ \varepsilon_{1_{\theta}}\\ \varepsilon_{1_{\nu}}\\\end{bmatrix} (k) \label{propagation formula}
    \end{align}
    Where $\vec{\varepsilon}_1$ is the error. To estimate the error, a normal distribution
    centered at 0 is assumed, where the covariances are our estimation of the maximum
    possible error (equation \ref{covariance w}). A diagonal covariance matrix
    is used since it is assumed that the errors in each internal state are
    uncorrelated.  The values given as covariance in the matrix $Q$ are estimated based on the specifications of the actuators.
    \begin{align}
        \begin{bmatrix}\varepsilon_{1_{x}}\\ \varepsilon_{1_{y}}\\ \varepsilon_{1_{\theta}}\\ \varepsilon_{1_{\nu}}\\\end{bmatrix} \simeq N(0,Q) &  & Q = Diag( \begin{bmatrix}0.2^{2}\\ 0.2^{2}\\ (3\frac{\pi}{180})^{2}\\ (0.5\frac{1000}{3600})^{2}\end{bmatrix} ) \label{covariance w}
    \end{align}
    Where $Diag(\vec{v})$ is a function that returns a square diagonal matrix with
    the elements of vector $\vec{v}$ on the main diagonal. With equations
    \ref{propagation formula} and \ref{covariance w} the propagation model
    defined in equation \ref{propagation model} is specified. To specify the measuring
    model (equation \ref{Measuring model}), we need to correlate the
    internal state with the outputs and add the error, as per equation \ref{Measuring
    formula}.

    \begin{equation}
        \textbf{y}(k) =
        \begin{bmatrix}
            x      \\
            y      \\
            \theta \\
            \nu    \\
            \omega
        \end{bmatrix}
        (k) =
        \begin{bmatrix}
            1 & 0 & 0 & 0                      \\
            0 & 1 & 0 & 0                      \\
            0 & 0 & 1 & 0                      \\
            0 & 0 & 0 & 1                      \\
            0 & 0 & 0 & \frac{\tan{\alpha}}{l}
        \end{bmatrix}
        \begin{bmatrix}
            x      \\
            y      \\
            \theta \\
            \nu    \\
        \end{bmatrix} (k)
        +
        \begin{bmatrix}
            \varepsilon_{2_{x}}      \\
            \varepsilon_{2_{y}}      \\
            \varepsilon_{2_{\theta}} \\
            \varepsilon_{2_{\nu}}    \\
            \varepsilon_{2_{\omega}}
        \end{bmatrix} (k)
        \label{Measuring formula}
    \end{equation}
    Where the $\vec{\varepsilon}_2$ error vector is approximated using a normal distribution,
    mean 0 and covariance the estimated maximum error, following equation \ref{covariance
    v}. In a similar fashion as equation \ref{covariance w} it is assumed that
    the errors are uncorrelated so a diagonal matrix is used. The values given as covariance in matrix $R$ are estimated based on the specifications of the sensors.
    \begin{align}
                \begin{bmatrix}
            \varepsilon_{2_{x}}\\ \varepsilon_{2_{y}}\\ \varepsilon_{2_{\theta}}\\ \varepsilon_{2_{\nu}}\\ \varepsilon_{2_{\omega}} \end{bmatrix}  \simeq N(0,R) &  & R = Diag( \begin{bmatrix}0.2^{2}\\ 0.2^{2}\\ (10\frac{\pi}{180})^{2}\\ (0.5\frac{1000}{3600})^{2}\\ (5\frac{\pi}{180}\frac{1}{60})^{2}\end{bmatrix} ) \label{covariance v}
    \end{align}
    Therefore, the model for the Extended Kalman filter is defined using
    equation \ref{propagation formula} with error \ref{covariance w} in the
    propagation step and equation \ref{Measuring formula} with error
    \ref{covariance v} for the measuring step.
    \subsubsection{Mapping}
\begin{figure}[!h]
        \centering
        \includegraphics[width=0.7\linewidth]{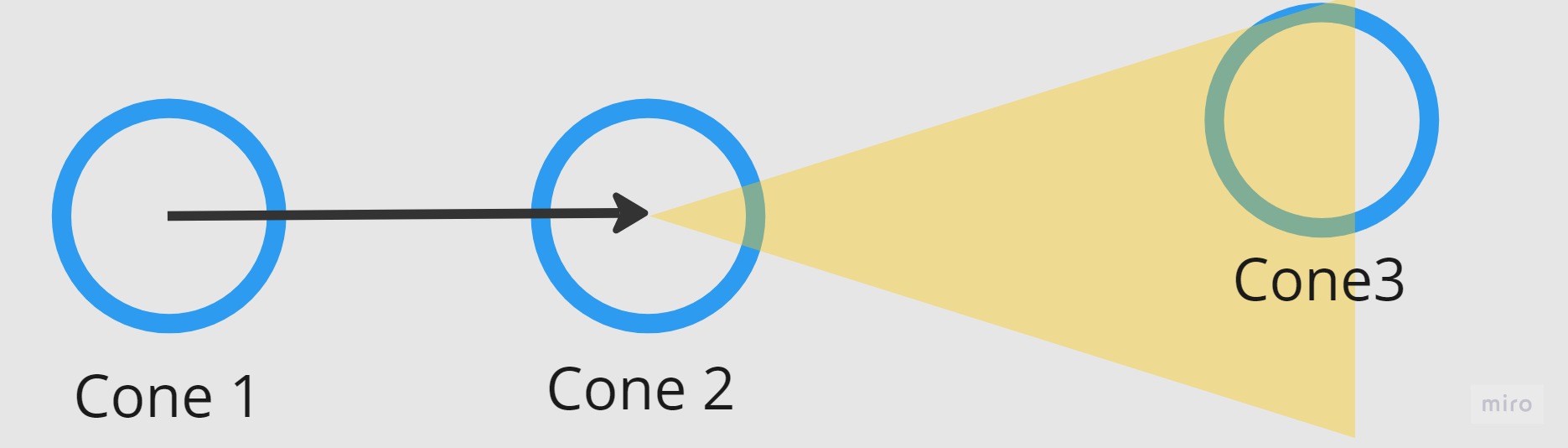}
        \caption{Cone ordering algorithm.}
        \label{cone_counting}
    \end{figure}

    The mapping node stores the cone and car positions across time. It uses the
    first values received by the Extended Kalman Filter as the world reference
    coordinates. From that point onward, all the coordinate information received
    will be transformed relative to that reference. The mapping node continuously
    receives the cone position and identity over time. To avoid the identiy
    switch errors, whenever the mapping node receives a new identity, it checks
    an oval area across the nearest cone. If the new detection is within this area,
    it is considered to be the same cone and it updates the new position to a
    weighted average of all the detections assigned to that cone. To calculate the
    average, the coefficient is inversely proportional to the distance, so the nearer
    cones weigh more than the farther ones. this mitigates the error generated
    by the depth camera. Since the search area changes inversely
    proportional with the distance, the nearest cones have a smaller search area
    than the farther ones. The mapping node publishes the cones currently visible,
    and all the cones detected so far. These cones must be ordered so the path-planning
    algorithm works properly. To order the cones an algorithm is used
    that searches in a triangular area following the vector defined by the
    previous cone of the same color as represented in figure \ref{cone_counting}.

    To define the beginning of either side of the track, the algorithm looks for
    the orange cones indicating the start line that are closest to the first blue/yellow
    cones of the track. The triangular search area varies depending on the
    mission. If several cones are detected in the area, the closest one to the
    previous cone will be the next counted one. If a detected cone does not fall
    in any search area, it is considered not relevant for this section of the track
    and will be ignored.
    \subsubsection{Map Awareness}
    The map awareness node uses the mapping and positioning information offered by
    the mapping node to calculate relevant information about the position of the
    car in the track. This information differs between missions. The information
    that this node provides is: the number of laps run, the current kind of lap (recon
    or timed), whether the finish line has been reached, if the vehicle is near the
    finish line, and if the vehicle is outside the track limits. In case of the skidpad
    mission, it also detects when the vehicle is near the skidpad central
    section, and the number of times it runs across it.
    \subsection{Path Planning}
    The Path Planning node calculates the position of the car at each moment, using the
    Spline submodule. The Spline module calculates the
    middle point of the track by interpolating the cones positions, and segments it into points
    with designated speeds for the Control node to follow.
    The racing line is computed as the midline between both track boundaries. First, the left and right track limits are defined by fitting independent 2D cubic splines through the ordered series of blue and yellow cones, respectively. These splines allow densely resampling each boundary at arbitrary resolution. Corresponding points on opposite splines are then paired to form transverse lines across the track, and the midpoint of each transverse line is interpolated to produce the racing line. This line is discretized into a sequence of waypoints for the Pure Pursuit controller to follow.
    The path planning algorithm operates in two modes. During the discovery lap, the spline is computed using only the currently visible cone detections. Once the system determines that the track forms a closed loop, global path planning activates and recomputes the spline over the entire mapped circuit \cite{adampathplanning2018}. For each waypoint along the discretized racing line, the node assigns a target speed constrained by the maximum allowed velocity, the maximum steering angle, and the steering rate, ensuring that the regulated Pure Pursuit controller can track the path at the assigned speeds without exceeding the vehicle's kinematic limits.
    
    \subsection{Global Map Alignment}
    For not closed circuit operations, it is necessary to position the ego vehicle
    against the defined track. In this kind of mission, it is insufficient to
    follow the track definition as it is discovered, specially if the track presents
    some kind of unusual layout definition. In this kind of
    layout, the pathplanning algorithm is not able to properly calculate a path
    following the mission rules. To fix these issues, we introduced an
    alternative pipeline that replaces mapping and pathplanning called Global Map Alignment.
    In this scenario, the position of the cones is known beforehand, so it is
    possible to store the position and path of the whole track and load it
    during the event. The Global Map Alignment node can accurately position the entire
    track by only seeing the first cones. It achieves this by storing a CSV file
    containing the position of all cones in the track, and mapping the coordinate
    frame given by positioning to ground truth using a 2x2 transformation matrix in order to cover rotation and scaling.

    When we first detect the orange cones in the beginning of the track, we calculate
    the transformation matrix:

    \begin{align*}
        T = \begin{bmatrix}a &b \\ c &d\end{bmatrix} &  & g = x_{c}y_{c}' - x_{c}' y_{c}           \\
        a = \frac{x_{t}y_{c}' - x_{t}' y_{c}}{g}     &  & b = \frac{x_{c}x_{t}' - x_{c}' x_{t}}{g} \\
        c = -\frac{y_{c}y_{t}' - y_{c}' y_{t}}{g}    &  & d = \frac{x_{c}y_{t}' - x_{c}' y_{t}}{g}
    \end{align*}

    Where $x_{c}$, $y_{c}$ are the coordinates of the cones from the vision module;
    and $x_{t}$, $y_{t}$ are the coordinates of the cones in the CSV track.

    This is just the basic transformation matrix; it does not necessarily transform all
    cones accurately, particularly because cones located farther from the start
    zone are highly sensitive to slight changes in the original input. To address
    this, we iteratively adjust the transformation matrix by using random
    numbers, aiming to minimize the following expression:

    \begin{align}
        \sum_{i=1}^{n}\frac{1}{d_{i}- d_{min}+ 1}
    \end{align}

    With $d$ defined as the distance from each cone to its predicted position. The
    fractional part ensures that cones that are closer are prioritized.

    \subsection{Control}
    The control module moves the vehicle along the calculated path using a Pure Pursuit
    algorithm. The Pure Pursuit Node calculates the target acceleration, steering
    and braking based on the path and speed profile from the path planning module,
    and the current car's position from the mapping module using a geometric
    following algorithm.

    The \textit{pure pursuit algorithm} works by defining a circular path that intersects
    with the spline given by path planning. This intersection defines a lookahead
    point, that is usually 3-5 seconds distance ahead. We define the acceleration at any given point as the difference between
    the current speed and the target speed (defined by the speed profile), multiplied
    by a constant factor of $0.5$ to avoid overshooting. Our implementation is a
    special case of \textit{pure pursuit algorithm} called \textit{regulated pure pursuit}. This algorithm differs
    from the general \textit{pure pursuit algorithm} by having a dynamic lookahead that can
    be parameterized. This lookahead value is then used in the following
    equation to determine the delta in steering:

    \begin{align}
        \alpha = t - \theta &  & \delta = \atantwo(2W * \sin(\alpha), l)
    \end{align}

    Where $t$ is defined as the direction to the target point; $\theta$, as the orientation
    of the car; $\delta$, as the change in direction needed; $W$, as the wheelbase
    of the vehicle; and $l$ as the selected lookahead.

    The lookahead is selected following the next equation:

    \begin{align}
        l = \max ( \min (L_{l}, |v_{t}- v|), L_{h})
    \end{align}

    Defining $L_{l}$ and $L_{h}$ as the lower and higher bounds of the lookahead;
    and $v_{t}$ and $v$ as the target speed as defined by the speed profile and
    the current speed.

    \subsection{Master Node}

    The master node is the system entrypoint. It will manage module activations
    and the system internal flags. To ensure that the system does not launch any
    unexpected actuator movements whenever the AS enters the \textit{driving}
    status, the master node activates the nodes from the path-planning and control
    packages. When the AS changes the status to any other, these nodes are deactivated.

    The master node checks the mission and the internal state variables provided
    via the CAN node, and sets a series of parameters to the other
    nodes. One of these parameters is the flag to set the initial coordinates and
    state vectors for the mathematical models.

    The master node checks the internal state of each of the nodes that form the
    AS. If any of these nodes enter an unexpected state, the master node activates
    the Emergency Brake System (EBS).

    \section{Results}

    The presented architecture was tested both as an end-to-end system, as well as
    the individual modules performance. Due to hardware accessibility
    restrictions, the end-to-end system was only tested in simulation, while the
    individual modules were tested with real world data. To avoid biases, the
    error distribution observed in the individual sensors during their
    individual evaluation was replicated in simulation to better reflect the
    global end-to-end system.

    \subsection{Object Detection}

    The FSOCO dataset was used to train the object detection model \cite{fsoco_2022}. This
    dataset provides over 10,000 annotated images for object detection. These annotation
    are high quality, since the dataset has quality assurance protocols, unlike the
    FSOCO Legacy dataset. The FSOCO Legacy contained a high number of very
    similar images since the teams usually provided images from the same video
    or the same track. The new FSOCO required that all the candidate dataset must
    have a low enough similarity score. Also, at least 50\% of the candidate
    data must be on-board footage of rules compliant tracks. The higher quality annotation
    and the low variability of the images entails a reduced training time.

    The dataset was divided between a 70\% training set, a 20\% validation set
    and a 10\% test set. The model was trained using the training set and the
    validation set was used to perform a hyperparameter search of the model. After
    the optimization, the test set was used to evaluate the overall performance.

\begin{figure}[H]
        \centering
        \includegraphics[width=\linewidth]{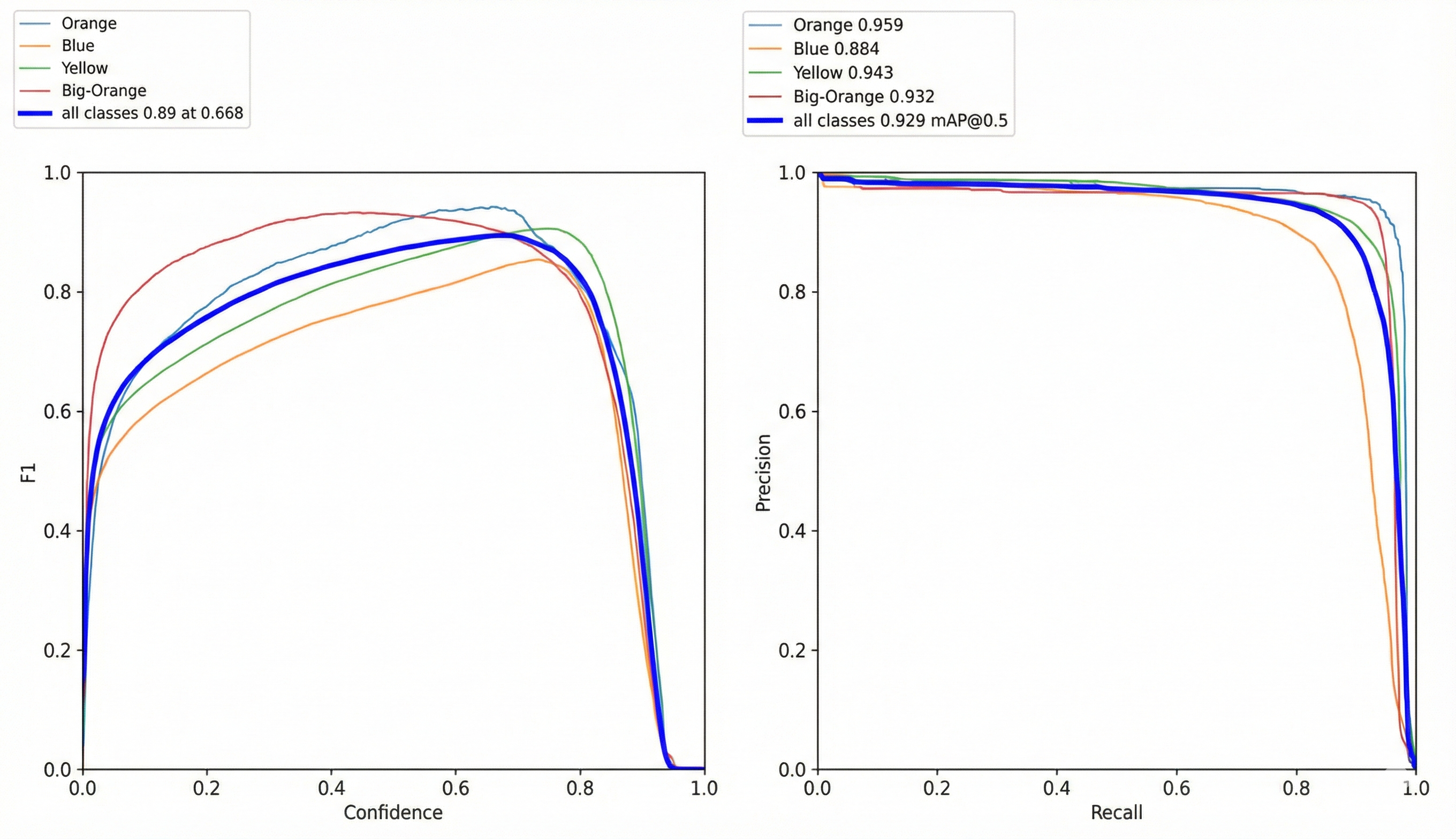}
        \caption{Object detection F1 curve and PR curve.}
        \label{fig:f1_curve}
    \end{figure}

\begin{figure}[H]
        \centering
        \includegraphics[width=\linewidth]{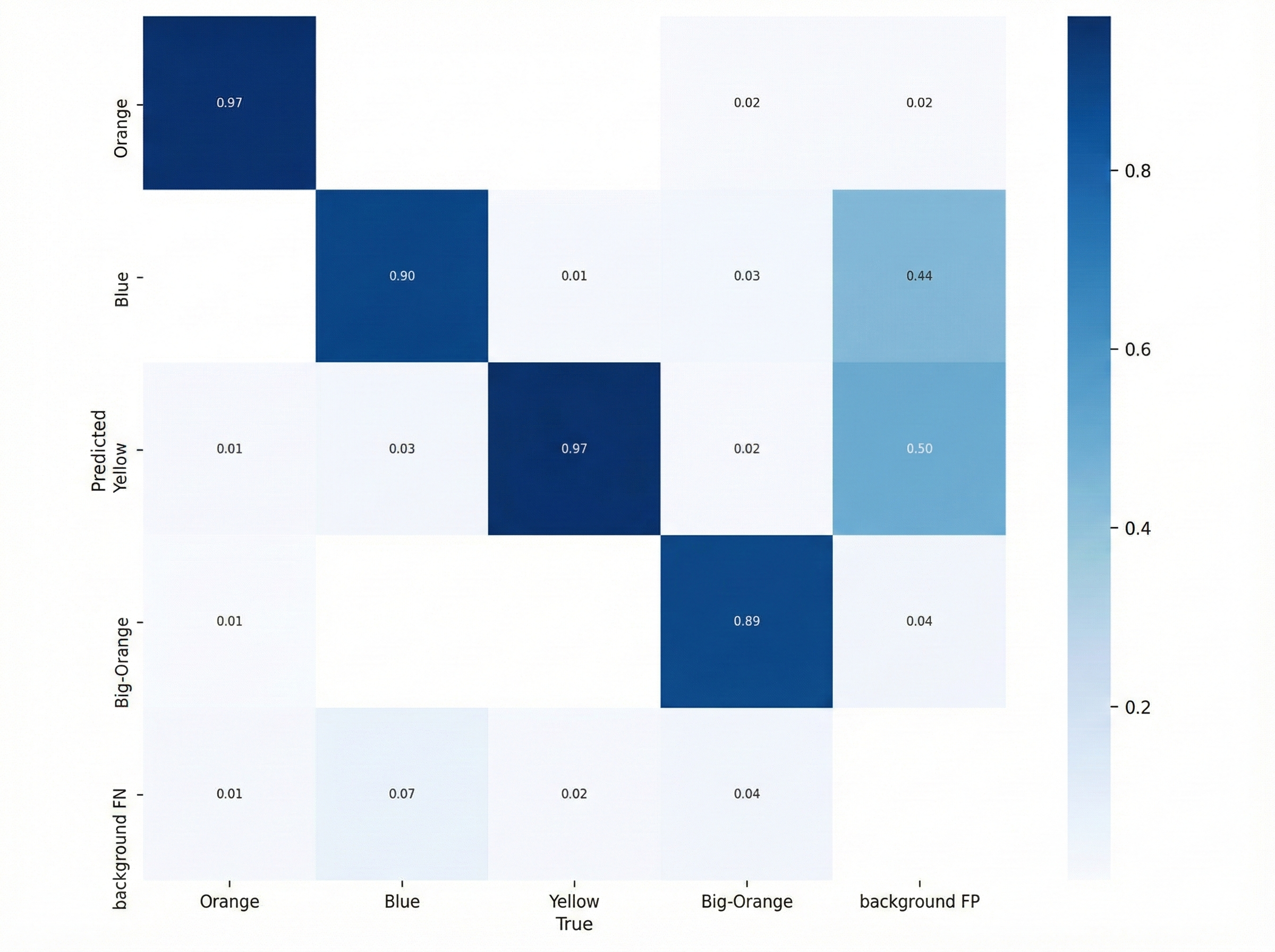}
        \caption{Object detection confusion matrix.}
        \label{fig:confusion_matrix}
    \end{figure}

    
    After optimizing the hyperparameters using the validation data, the model was evaluated in its classification
    capabilities using the test data. The results are shown in Figures \ref{fig:f1_curve}
    and \ref{fig:confusion_matrix}.

    \subsection{Depth calculation}
    
\begin{figure}[H]
        \centering
        \includegraphics[width=\linewidth]{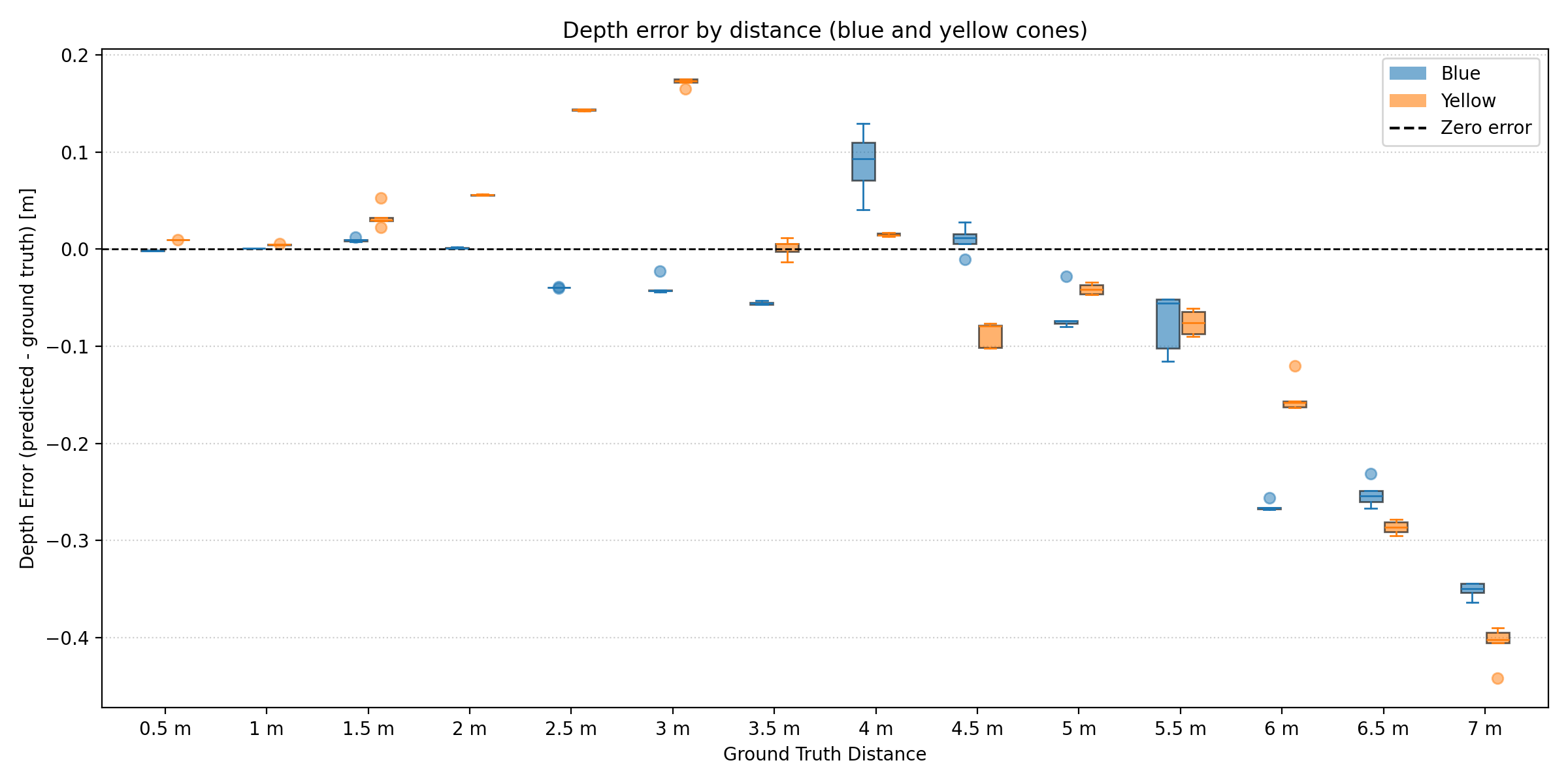}
        \caption{Depth Camera error distribution}
        \label{Depth_error}
    \end{figure}

    To evaluate the depth calculation algorithms, a real-world testing was done.
    For this process, the cones that are the detection objectives were located at
    increasing distances from the camera, and the distance estimation was
    measured comparatively to the real distance. This test was done at daylight to
    better evaluate the performance in a real environment. The result of this
    testing can be seen in Figure \ref{Depth_error}

   As shown in Figure \ref{Depth_error}, the depth estimation error increases approximately exponentially with distance, with a systematic bias toward underestimating range (placing objects closer than their true position). The median error reaches 0.5 m at a ground-truth distance of 7 m. A color-dependent discrepancy is also observed: blue cones are consistently estimated as closer than yellow cones, particularly at medium distances around 3 m, likely due to differences in how the neural depth model responds to each color under natural lighting. Despite this, the error margins remain acceptable for the application, as the systematic underestimation introduces a conservative bias — the vehicle initiates maneuvers earlier than strictly necessary, providing an implicit safety margin in real-world operation.

    \subsection{RTK data}

\begin{figure}[H]
        \centering
        \includegraphics[width=0.8\linewidth]{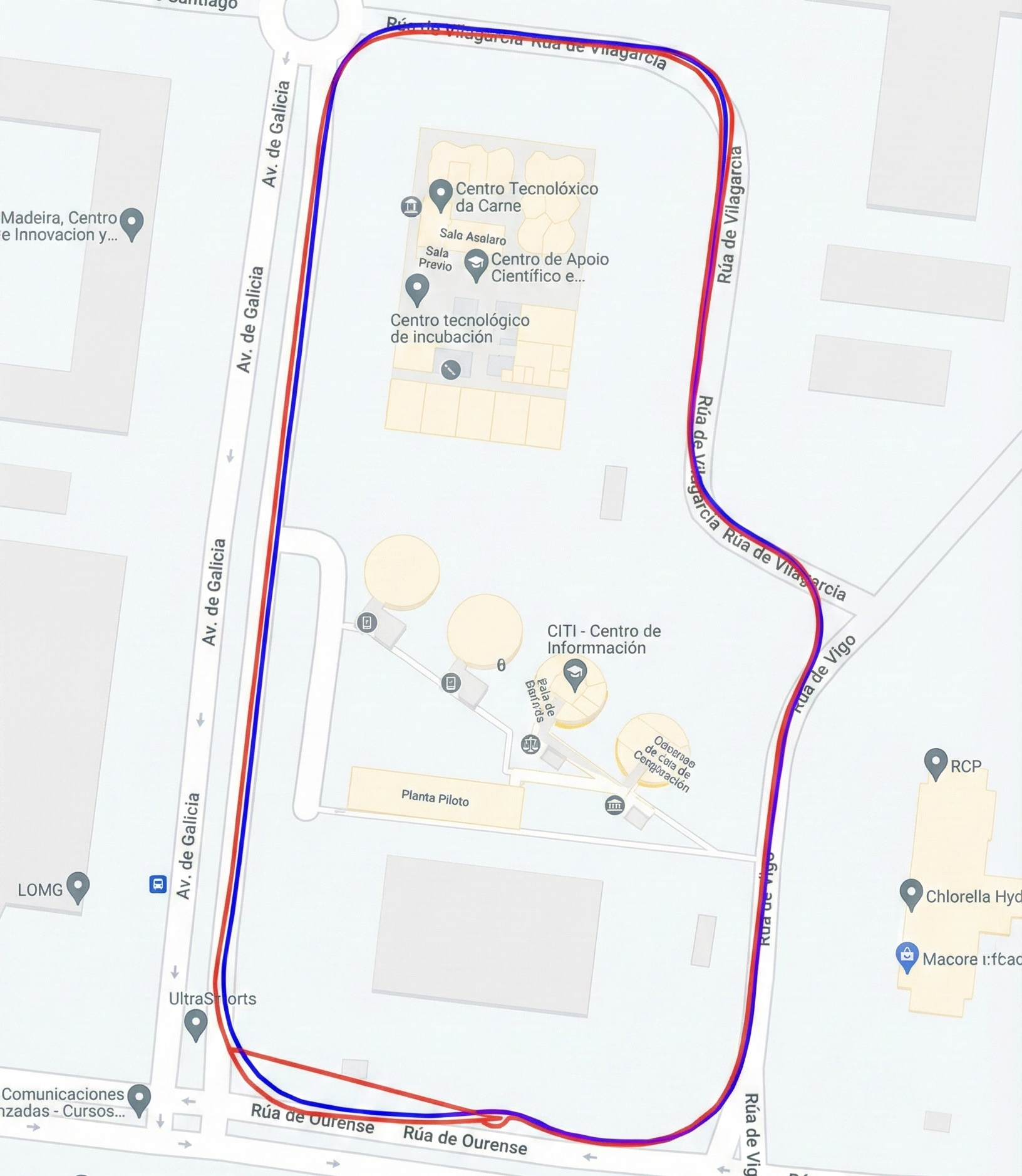}
        \caption{GNSS (Red) vs RTK (Blue) comparative}
        \label{fig:circuit_rtk}
    \end{figure}

\begin{figure}[H]
        \centering
        \includegraphics[width=\linewidth]{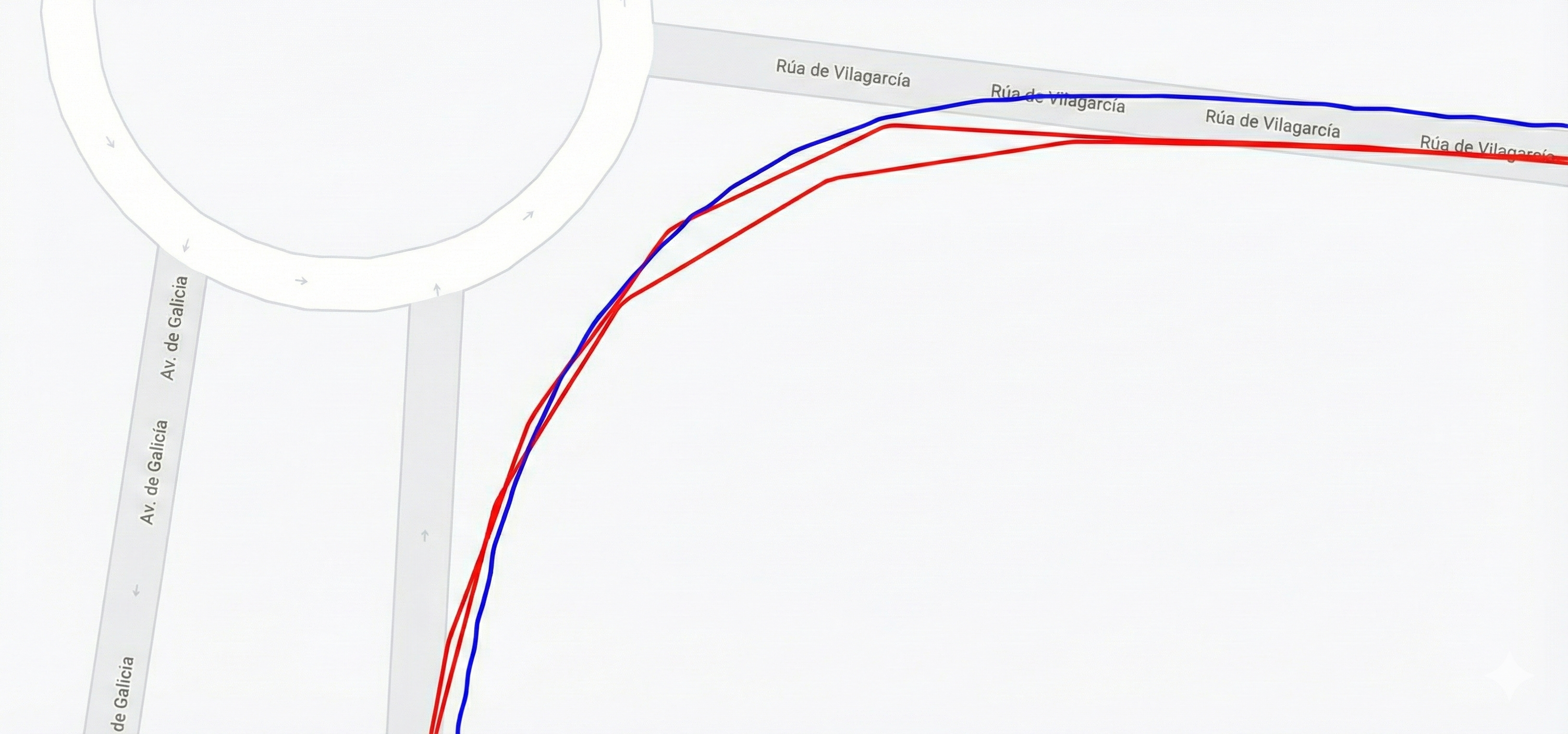}
        \caption{Zoomed in section of the circuit.}
        \label{fig:circuit_zoomed}
    \end{figure}

    The RTK data was evaluated comparatively with the GNSS data. This was done
    via a closed circuit track and measuring both the raw GNSS and the corrected
    coordinates via RTK. The measurements can be seen in Figures \ref{fig:circuit_rtk}
    and \ref{fig:circuit_zoomed}.

    As it can be seen, the RTK provides more accurate data, and the loss of signal
    is more infrequent. These losses of signal in GNSS can be detected in the "straight
    lines" that are shown on the graph. In the measurements done in this experiments,
    the precision improved up to a 12 centimeter error using RTK.

    \subsection{Car Controls}

\begin{figure}[H]
        \centering
        \includegraphics[width=\linewidth]{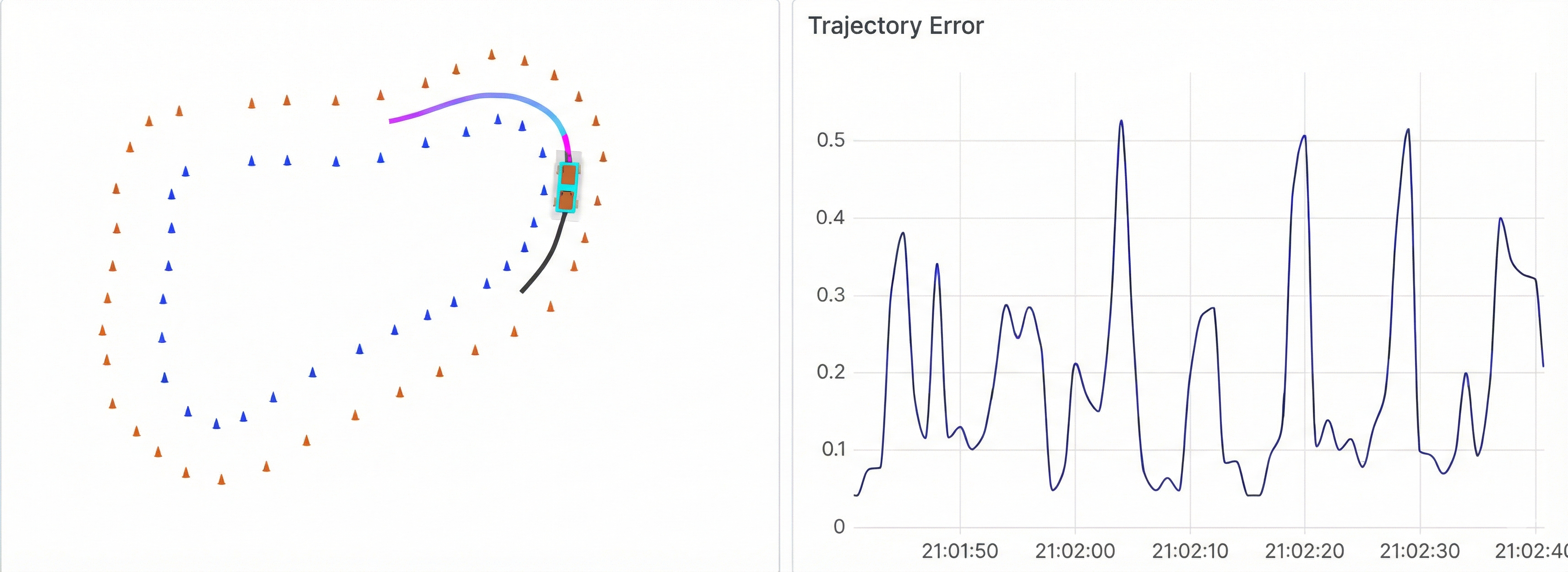}
        \caption{Track positioning and error graph.}
        \label{fig:control_error}
    \end{figure}

    Finally, to evaluate the car control algorithms, the pipeline is deployed in
    a simulation environment. In this environment the track is deployed and the
    perception part of the pipeline is simulated, incorporating the error distribution
    perceived in the previous experiments. A segment of the circuit is
    illustrated in figure \ref{fig:control_error}.

    \clearpage

    \section*{Acknowledgments}

    This project is supported from the Escola Superior de Enxeñería Informática and
    Escola Técnica Superior de Enxeñería Industrial from Universidade de Vigo.
    Especially with direct help from professors Dr. Javier Rodeiro Iglesias, Dr. Enrique Paz Domonte and Dr. Joaquín López Fernández.

    \printbibliography
\end{document}